\documentclass[conference]{IEEEtran}
\IEEEoverridecommandlockouts
\usepackage{cite}
\usepackage{amsmath,amssymb,amsfonts}
\usepackage{algorithmic}
\usepackage{graphicx}
\usepackage{textcomp}
\usepackage{xcolor}
\usepackage{booktabs}
\usepackage{url}
\usepackage{float}
\usepackage{graphicx} 
\usepackage{subcaption} 
\def\BibTeX{{\rm B\kern-.05em{\sc i\kern-.025em b}\kern-.08em
    T\kern-.1667em\lower.7ex\hbox{E}\kern-.125emX}}

\begin{document}

\title{DynoSLAM: Dynamic SLAM with Generative Graph Neural Networks for Real-World Social Navigation}


\author{
\IEEEauthorblockN{
Danil Tokhchukov\textsuperscript{1}, 
Veronika Morozova\textsuperscript{1}, 
Gonzalo Ferrer\textsuperscript{1}
}
\thanks{\textsuperscript{1} Applied AI Institute, Moscow, Russia.}
\thanks{Code is available at \protect\url{https://github.com/makriot/dynoslam}}
}

\maketitle

\begin{abstract}

Traditional Simultaneous Localization and Mapping (SLAM) algorithms rely heavily on the static environment assumption, which severely limits their applicability in real-world spaces populated by moving entities, such as pedestrians. In this work, we propose DynoSLAM, a tightly-coupled Dynamic GraphSLAM architecture that integrates socially-aware Graph Neural Networks (GNNs) directly into the factor graph optimization. Unlike conventional approaches that use rigid constant-velocity heuristics or deterministic single-agent neural priors, our framework formulates pedestrian motion forecasting as a stochastic World Model. By utilizing Monte Carlo rollouts from a trained GNN, we capture the multimodal epistemic uncertainty of human interactions and embed it into the SLAM graph via a dynamic Mahalanobis distance factor. We demonstrate through extensive simulated experiments that this stochastic formulation not only maintains highly accurate retrospective tracking (Robot RMSE of 0.39m) but also prevents the optimization failures caused by the deterministic "argmax problem". Ultimately, extracting the empirical mean and covariance matrices of future pedestrian states provides a mathematically rigorous, probabilistic safety envelope for downstream local planners (e.g., MPC or PPO), enabling anticipatory and collision-free robot navigation in densely crowded environments.
\end{abstract}




\section{Introduction}
The fundamental premise of conventional GraphSLAM systems is the rigidity of the environment. When navigating through crowded, human-centric spaces, this static assumption is severely violated. Dynamic landmarks introduce false observation factors that corrupt the optimization process, leading to severe trajectory drift or complete localization failure. 

Recent advancements attempt to mitigate this by masking out dynamic features or tracking moving objects using simple kinematic heuristics. For instance, recent works in 3D Dynamic Scene Graphs (DSG) \cite{dsg_slam} and Gaussian Splatting \cite{dynagslam} model dynamic entities using constraint-based formulations, such as Constant Velocity Models (CVM). While mathematically simple, these linear heuristics completely fail to anticipate the complex, non-linear avoidance maneuvers characteristic of human crowds. Attempts to replace these heuristics with standard deterministic neural networks often suffer from the ``argmax problem'': when faced with the inherent ambiguity of social interactions, deterministic models either collapse to unphysical straight lines or produce highly jittery trajectories that over-constrain and misdirect the SLAM optimizer.

In this work, we introduce a paradigm shift with \textbf{DynoSLAM}: instead of relying on hardcoded physics or rigid deterministic predictions, we augment the GraphSLAM factor graph with socially-aware, stochastic motion priors. By employing a Graph Attention Network (GAT) as a lightweight World Model and querying it via Monte Carlo rollouts, we explicitly sample the multimodal distribution of human intents. 

This stochastic formulation allows us to extract not just the expected future trajectories of pedestrians, but also their corresponding epistemic uncertainty (covariance). By embedding these statistics into the SLAM optimization via a dynamic Mahalanobis distance factor, the system automatically adjusts the stiffness of the kinematic prior—tightly binding the graph in unambiguous situations and relaxing it during complex, unpredictable interactions. Ultimately, DynoSLAM bridges the gap between geometric state estimation and context-aware motion forecasting, providing downstream control algorithms (such as MPC or PPO) with the mathematically rigorous probabilistic safety envelopes required for safe, anticipatory robot navigation.




\section{Related Work}

\subsection{Dynamic SLAM and Object Tracking}
Dynamic SLAM approaches aim to jointly estimate the camera/robot trajectory and the states of moving objects in the environment. VDO-SLAM \cite{vdo_slam} extracts dynamic SE(3) poses but relies on decoupled motion models that do not feed tracking uncertainty back into the optimization. A highly relevant state-of-the-art framework, DynaGSLAM \cite{dynagslam}, performs online tracking and motion prediction of dynamic 3D Gaussians. However, it relies on a simplistic Constant Acceleration model, inherently ignoring the contextual intentions of agents and failing during non-linear avoidance maneuvers. Similarly, recent DSG-SLAM architectures \cite{dsg_slam} tightly couple semantic scene generation with dynamic object tracking using rigid body constraints. While geometrically robust, these methods lack the probabilistic generative priors necessary for complex social trajectory forecasting, treating all predicted futures with equal (and often misplaced) confidence.

\subsection{Social Motion Prediction and the Argmax Problem}
Predicting human trajectories in crowded spaces is a well-studied problem, typically addressed using Recurrent Neural Networks (RNNs) or Graph Neural Networks (GNNs). While these data-driven models capture social context, integrating them directly into a SLAM factor graph or a control loop is challenging. Standard deterministic networks trained with Mean Squared Error (MSE) losses suffer from the ``argmax problem'': when forced to output a single trajectory in an inherently ambiguous situation (e.g., passing a pedestrian on the left versus the right), the model often averages the possible outcomes, producing unphysical, collision-prone straight lines. Generative models, such as Social-GAN or diffusion-based predictors, can generate multimodal future hypotheses, but their integration into real-time factor graphs remains underexplored. Our work bridges this gap by interpreting a stochastic GNN as a lightweight World Model, using Monte Carlo rollouts to compute dynamic covariance matrices directly within the SLAM objective.

\subsection{Uncertainty-Aware Human Avoidance}
Safe robotic navigation in dynamic environments requires not only current state estimation but also a quantified assessment of future trajectory uncertainty. Approaches based on Model Predictive Control (MPC), Model Predictive Path Integral (MPPI), and Proximal Policy Optimization (PPO) \cite{smpc, rl_navigation} are widely used for human avoidance. However, these controllers typically assume access to perfect, noise-free future trajectories from decoupled perception modules, or rely on artificially inflated bounding boxes. Our approach differs by feeding the controller with exact mean trajectories and dynamically expanding covariance ellipses that are jointly optimized within the SLAM backend. This ensures mathematically rigorous, uncertainty-aware navigation where the robot reacts to the epistemic ambiguity of human intent rather than blindly trusting a single predicted path.

\section{Method}

\subsection{Standard GraphSLAM vs. Proposed Dynamic Formulation}
Let $X = \{x_1, \dots, x_T\}$ represent the robot trajectory, $U = \{u_1, \dots, u_T\}$ the odometry inputs, and $M = \{m_1, \dots, m_K\}$ a set of landmarks. In standard static SLAM, landmarks are assumed to be stationary (time-invariant). Given a set of observations $Z$, where $z_{j,i}$ denotes the measurement associated with landmark $j$ at time step $i$, classic GraphSLAM optimizes the Maximum A Posteriori (MAP) objective:
\begin{equation} \label{eq:standard_slam}
\begin{aligned}
    J_{static}(X, M) &= \sum_{i=1}^T \Big( \|x_i - g(x_{i-1}, u_i)\|_{\Sigma_{u}}^2 \\
    &\quad + \sum_{j \in \mathcal{V}_i} \|z_{j,i} - h(x_i, m_j)\|_{\Sigma_{z}}^2 \Big)
\end{aligned}
\end{equation}
where $g(\cdot)$ is the robot kinematic model, $h(\cdot)$ is the sensor observation model, and $\mathcal{V}_i$ is the subset of landmarks visible to the robot at time $i$ (data association mapping). The norms represent Mahalanobis distances weighted by the inverse covariance matrices $\Sigma_u^{-1}$ and $\Sigma_z^{-1}$.

In dynamic environments, the stationary assumption $m_j = \text{const}$ fails. To model moving entities, we introduce time-dependency for the landmarks $m_{j,i}$ and define an augmented joint state vector $y_i = [x_i, m_{1,i}, \dots, m_{K,i}]^T$. The system's temporal evolution is now governed by a joint transition function $p(y_i \mid y_{i-1}, u_i)$, which updates both the robot pose and the dynamic landmarks simultaneously:
\begin{equation}
    y_i =
    \begin{bmatrix}
    g(x_{i-1}, u_i) \\
    m_{1,i-1} + v_{1,i-1} \cdot \delta t \\
    \vdots \\
    m_{K,i-1} + v_{K,i-1} \cdot \delta t
    \end{bmatrix} + w_i
\end{equation}
where $v_{j,i-1}$ is the velocity of the $j$-th landmark and $w_i$ is the combined process noise.

Consequently, we reformulate the GraphSLAM objective to optimize the joint state sequences $Y = \{y_1, \dots, y_T\}$. By decoupling the joint transition into the robot's odometry factor and object-specific kinematic factors $\mathcal{L}_{kin}$, the proposed dynamic objective becomes:
\begin{equation} \label{eq:dynamic_slam}
\begin{aligned}
    J_{dynamic}(Y) &= \sum_{i=1}^T \Big( \|x_i - g(x_{i-1}, u_i)\|_{\Sigma_{u}}^2 \\
    &\quad + \sum_{j \in \mathcal{V}_i} \|z_{j,i} - h(x_i, m_{j,i})\|_{\Sigma_{z}}^2 \\
    &\quad + \sum_{j=1}^K \mathcal{L}_{kin}^{(j,i)} \Big)
\end{aligned}
\end{equation}
Here, $\mathcal{L}_{kin}^{(j,i)}$ acts as a soft kinematic prior factor governing the physical evolution of the dynamic object, replacing hardcoded constant velocity assumptions with a data-driven distribution.

\subsection{Landmark History Formulation}
To predict the evolution of a dynamic landmark, our models rely on its spatial history $\mathcal{H}_{k,i}$. 
During \textbf{offline training}, $\mathcal{H}_{k,i}$ is extracted directly from the ground-truth trajectories (e.g., bounding box centroids from autonomous driving datasets). 
During \textbf{online SLAM inference}, the history is constructed dynamically from the previously optimized states within the GraphSLAM solver's active sliding window:
\begin{equation}
    \mathcal{H}_{k,i} = \{ m_{k, i-H}^*, m_{k, i-H+1}^*, \dots, m_{k, i-1}^* \}
\end{equation}
where $H$ is the history length and $m^*$ denotes the posterior MAP estimates computed in the previous optimization steps.

\subsection{Constant Velocities (Baseline)}
As a foundational non-neural baseline, we model $\mathcal{L}_{kin}$ by adding explicit velocity nodes $v_{k,i}$ to the graph and penalizing acceleration (assuming a constant velocity transition):
\begin{equation}
    \mathcal{L}_{kin}^{(k,i)} = \| m_{k,i} - m_{k,i-1} - v_{k,i-1}\delta t \|_{\Sigma_p}^2 + \| v_{k,i} - v_{k,i-1} \|_{\Sigma_v}^2
\end{equation}

Alternatively, by substituting the velocity definitions $v_{k,i-1} = \frac{m_{k,i} - m_{k,i-1}}{\delta t}$ and $v_{k,i} = \frac{m_{k,i+1} - m_{k,i}}{\delta t}$, the constant velocity factor can be expressed solely in terms of landmark positions, removing the need for explicit velocity nodes:
\begin{equation}
\mathcal{L}_{kin}^{(k,i)} = \frac{1}{(\delta t)^2} \left\| m_{k,i+1} - 2m_{k,i} + m_{k,i-1} \right\|_{\Sigma_v}^2.
\end{equation}

\subsection{Single Landmark Velocity Prediction}
To capture non-linear motion, we replace the hardcoded baseline with a neural network (e.g., an MLP or RNN) that predicts the next velocity $\hat{v}_{k,i} = s_\theta(\mathcal{H}_{k,i})$. While this data-driven prior eliminates the need for explicit velocity nodes in the graph, it treats each agent independently, leading to potential collision predictions in crowded spaces.

\textbf{1. Context Graph Construction and Social Attention:}
For a target landmark $k$ at time step $i-1$, we identify its local neighborhood $\mathcal{N}(k)$ using an $L_2$ spatial window of radius $R$:
\begin{equation}
    \mathcal{N}(k) = \left\{ j \mid \| m_{k,i-1} - m_{j,i-1} \|_2 \leq R \right\}.
\end{equation}
For each pedestrian, a short trajectory history $\mathcal{H}_{k,i}$ is encoded into a latent feature $h_k = f_{\text{hist}}(\mathcal{H}_{k,i})$. We then construct a graph attention mechanism over neighboring pedestrians in order to model pairwise social interactions. The attention coefficient $\alpha_{k,j}$ quantifies the influence of neighbor $j$ on the future motion of pedestrian $k$:
\begin{equation}
    \alpha_{k,j} =
    \frac{
        \exp\left(
            \text{LeakyReLU}\left(
                a^\top [W h_k \parallel W h_j]
            \right)
        \right)
    }{
        \sum_{n \in \mathcal{N}(k)}
        \exp\left(
            \text{LeakyReLU}\left(
                a^\top [W h_k \parallel W h_n]
            \right)
        \right)
    }.
\end{equation}
The corresponding social context vector is obtained as
\begin{equation}
    c_k = \sum_{j \in \mathcal{N}(k)} \alpha_{k,j} W h_j.
\end{equation}
To preserve the pedestrian's own inertial motion, we concatenate the individual history feature and the aggregated social context, forming a joint representation
\begin{equation}
    z_k = [h_k \parallel c_k],
\end{equation}
which is then used to predict the next-step velocity.

\textbf{2. Deterministic Autoregressive GAT Prediction:}
Our first predictor is a deterministic GAT-based velocity model. Given the current history and the social context, the network outputs the next velocity estimate
\begin{equation}
    \hat{v}_{k,i} = g_\theta(z_k),
\end{equation}
where $g_\theta(\cdot)$ is a learnable MLP head on top of the GAT representation. The next position is then obtained by Euler integration:
\begin{equation}
    \hat{m}_{k,i} = \hat{m}_{k,i-1} + \hat{v}_{k,i} \Delta t.
\end{equation}
To predict multiple future steps, we use the model autoregressively: the newly predicted position is appended to the trajectory history, the oldest history element is discarded, and the graph attention is recomputed at the next step. Repeating this procedure yields a future rollout
\begin{equation}
    \hat{\tau}_k = \left\{ \hat{m}_{k,i}, \hat{m}_{k,i+1}, \dots, \hat{m}_{k,i+T} \right\}.
\end{equation}

\textbf{3. Stochastic GAT via Monte Carlo Rollouts:}
In practice, the deterministic GAT often produces future trajectories that are locally plausible but noticeably diverse and sometimes jittery across nearby interaction scenarios. Instead of forcing GraphSLAM to trust a single rollout, we reinterpret this variability as uncertainty in the human motion prior.

To obtain a stochastic predictor without retraining the backbone, we use the same trained GAT weights and perform multiple autoregressive rollouts with small Gaussian perturbations injected into the predicted velocity at each step:
\begin{equation}
    \tilde{v}_{k,i}^{(s)} = g_\theta(z_k^{(s)}) + \varepsilon_{k,i}^{(s)},
    \qquad
    \varepsilon_{k,i}^{(s)} \sim \mathcal{N}(0, \sigma_{\text{sto}}^2 I),
\end{equation}
where $s \in \{1, \dots, N\}$ indexes the sampled trajectory. Each sampled velocity is integrated autoregressively:
\begin{equation}
    \tilde{m}_{k,i}^{(s)} = \tilde{m}_{k,i-1}^{(s)} + \tilde{v}_{k,i}^{(s)} \Delta t.
\end{equation}
Because the perturbed rollout is fed back into the history buffer, the interaction graph and attention weights are recomputed separately for each sample, producing a set of socially-aware future hypotheses
\begin{equation}
    \left\{ \tilde{\tau}_k^{(1)}, \dots, \tilde{\tau}_k^{(N)} \right\}.
\end{equation}

\textbf{4. Mean and Uncertainty Estimation:}
The stochastic GAT allows us to estimate both the expected future velocity and its uncertainty. From $N$ sampled velocity hypotheses at step $i$, we compute the empirical mean
\begin{equation}
    \mu_{k,i} = \frac{1}{N} \sum_{s=1}^{N} \tilde{v}_{k,i}^{(s)}
\end{equation}
and the empirical covariance
\begin{equation}
    \Sigma_{k,i} =
    \frac{1}{N-1}
    \sum_{s=1}^{N}
    \left(\tilde{v}_{k,i}^{(s)} - \mu_{k,i}\right)
    \left(\tilde{v}_{k,i}^{(s)} - \mu_{k,i}\right)^\top.
\end{equation}
Intuitively, when pedestrian motion is simple and unambiguous, the samples remain concentrated and $\Sigma_{k,i}$ is small. In crowded or interaction-heavy scenes, the rollout set spreads out, yielding a larger covariance and therefore a weaker kinematic prior.

\subsection{Integration into GraphSLAM}
The predicted motion prior is incorporated into the dynamic GraphSLAM backend through a kinematic factor applied to each landmark trajectory.

\textbf{Variant 1 (Deterministic GAT, $L_2$ Loss):}
If only a single deterministic prediction is used, the kinematic factor penalizes deviation from the predicted displacement:
\begin{equation}
    \mathcal{L}_{kin}^{(k,i)} =
    \left\|
        m_{k,i} - m_{k,i-1} - \hat{v}_{k,i}\Delta t
    \right\|_2^2.
\end{equation}
This formulation is simple, but it treats all predictions as equally reliable and may over-constrain the optimizer when the predicted trajectory is uncertain or multimodal.

\textbf{Variant 2 (Stochastic GAT, Mahalanobis Loss):}
Using the sampled stochastic rollouts, we replace the single velocity hypothesis by its empirical mean $\mu_{k,i}$ and covariance $\Sigma_{k,i}$. The corresponding factor is written as a Mahalanobis distance:
\begin{equation}
    \mathcal{L}_{kin}^{(k,i)} =
    \left\|
        m_{k,i} - m_{k,i-1} - \mu_{k,i}\Delta t
    \right\|_{\Sigma_{k,i}^{-1}}^2.
\end{equation}
This formulation dynamically modulates the stiffness of the kinematic factor. If the stochastic GAT is confident, the covariance is small and the prior becomes strong. If the predicted futures are diverse, the covariance grows and the optimizer automatically down-weights the motion prior, relying more on observations.


\subsection{Application to Downstream Planning}
The optimized landmark states together with the stochastic motion statistics $\{\mu_{k,i}, \Sigma_{k,i}\}$ can be streamed to a downstream planner. In particular, MPC or PPO-based controllers may use $\Sigma_{k,i}$ to construct uncertainty-aware safety margins around pedestrians. This enables the robot to perform anticipatory and collision-aware navigation directly from onboard perception and SLAM, without requiring an external motion capture system.

\section{Experiments}
\label{sec:experiments}

The proposed pipeline is evaluated on a custom human avoidance benchmark within a simulated 2D environment. Our experimental roadmap is designed to incrementally demonstrate the necessity of socially-aware and stochastic motion priors in dynamic SLAM. We analyze the performance of static baselines, the Constant Velocity Model (CVM), single-agent neural predictors (MLP), and finally our proposed multi-agent approaches (Deterministic and Stochastic GAT).

\subsection{Setup and Environment}
\label{subsec:setup}
To construct a realistic and highly interactive pedestrian environment, we utilized \textit{pyminisim} \cite{pyminisim}, a 2D social navigation simulator. The movement of pedestrians in the simulator is governed by the Headed Social Force Model (HSFM) \cite{farina2017walking}, which generates complex, non-linear avoidance trajectories by accounting for the pedestrians' heading directions and social interactions. The robot navigation is controlled using a Model Predictive Control (MPC) algorithm.

Using this environment, we collected a large-scale dynamic SLAM dataset comprising 6,000 distinct episodes, with number of pedastrians varying from 1 to 15. Each episode contains continuous trajectories of the robot and multiple interacting pedestrians, alongside noisy odometry and local range-bearing observations. We partitioned the dataset into 5,500 episodes for training the neural velocity predictors (MLP, GAT) and 500 episodes for testing the full SLAM pipeline.

\begin{figure}[ht]
    \centering
    \includegraphics[width=\linewidth]{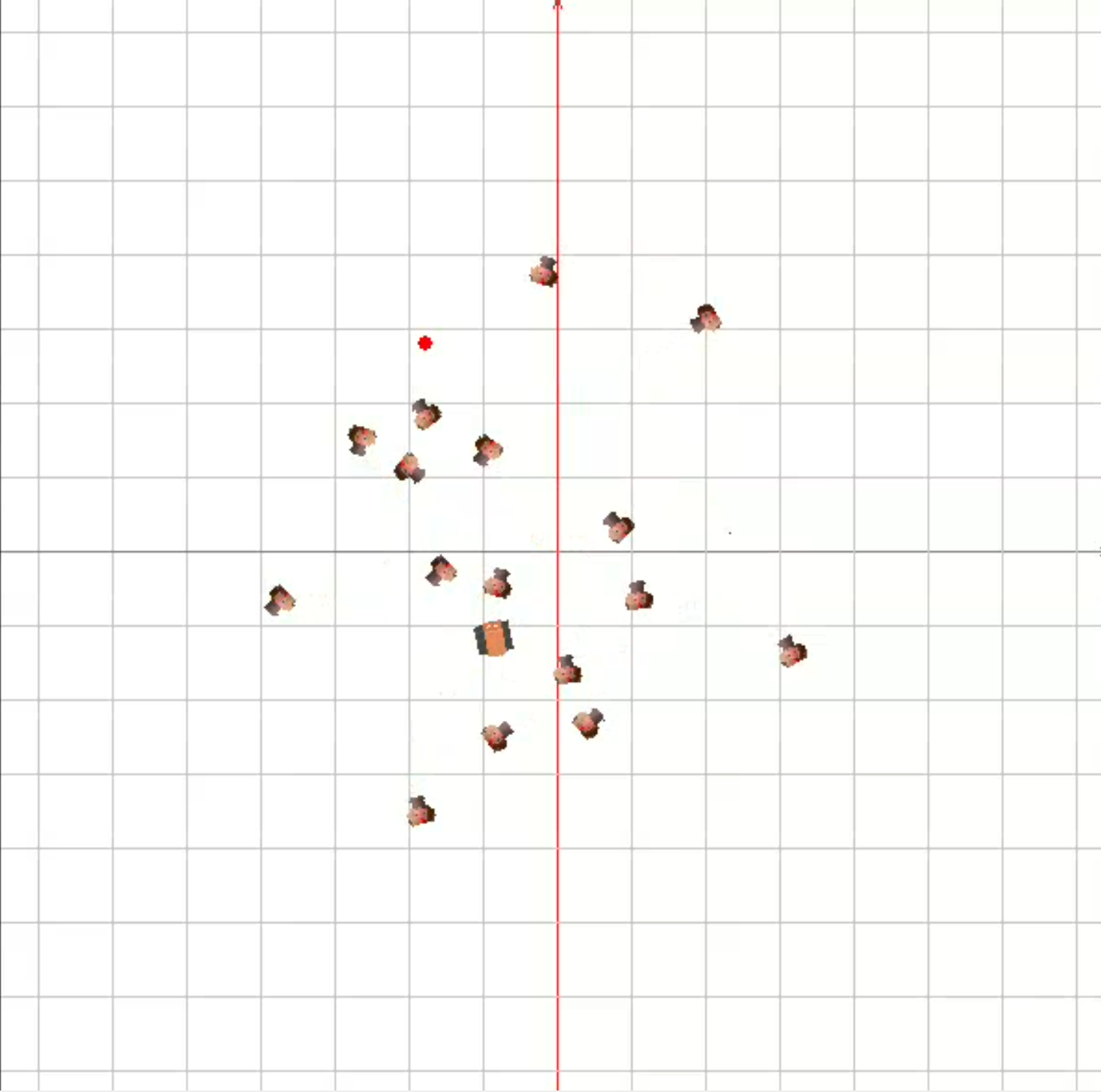}
    \caption{A screenshot of the \textit{pyminisim} 2D social navigation simulator. Pedestrians (controlled by HSFM) dynamically interact and avoid each other, while the robot (controlled by MPC) navigates through the crowd.}
    \label{fig:simulator}
\end{figure}



\subsection{Ablation of Kinematic Prior and Constant Velocity Baseline}
\label{subsec:exp_cvm}
First, we evaluate the baseline performance of GraphSLAM using the Constant Velocity Model (CVM) as the kinematic factor. In this setup, the prior mathematically assumes that the velocity of each pedestrian remains constant over time. Additionally, we conduct a structural ablation by completely removing the kinematic loss ($\mathcal{L}_{kin} = 0$), forcing the optimizer to rely strictly on noisy sensor observations.

As shown in Table~\ref{tab:metrics_evaluation}, the retrospective localization performance (Robot RMSE, LM RMSE, ATE, and RPE) remains largely consistent across the methods, including the $\mathcal{L}_{kin} = 0$ ablation. This indicates that within the optimization window, dense sensor observations dominate the factor graph, effectively constraining the historical trajectory. However, the critical failure of the $\mathcal{L}_{kin} = 0$ setup is its inability to extrapolate; without a kinematic factor to estimate velocities, future trajectory prediction (Prediction RMSE) becomes completely impossible.

\begin{figure}[ht]
    \centering
    \includegraphics[width=\linewidth]{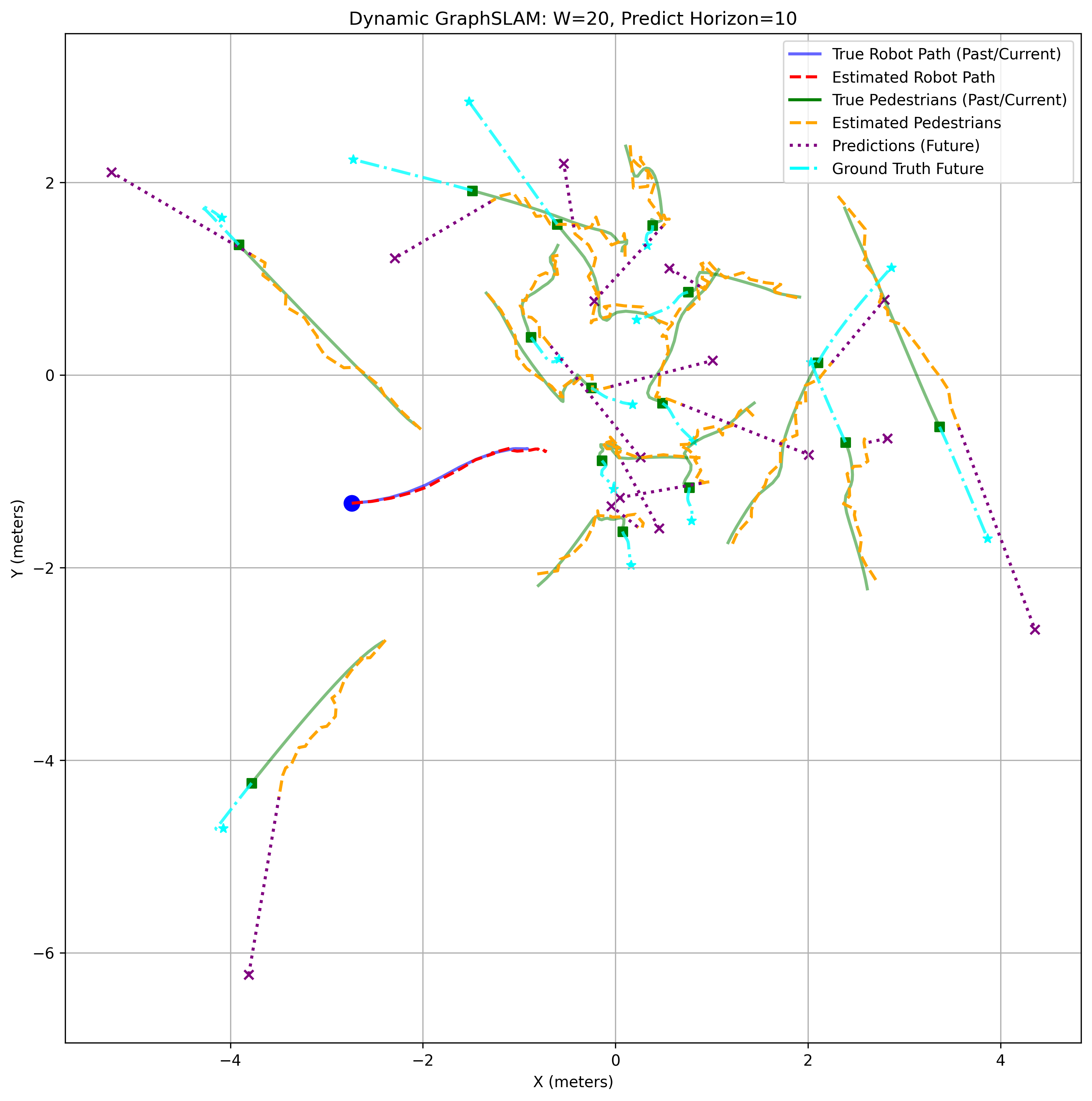}
    \caption{CVM Baseline Predictions. The retrospective past trajectories are well-optimized thanks to dense observations, but the predicted future path (purple) strictly follows a straight line, failing to capture the true non-linear avoidance maneuver (cyan).}
    \label{fig:cvm_baseline}
\end{figure}

Incorporating the CVM prior solves this by bridging retrospective optimization with future forecasting, establishing a baseline Prediction RMSE of 1.210. While CVM allows the system to look ahead, its inherent linearity causes it to fail during sudden avoidance maneuvers. Because CVM strictly extrapolates the last known velocity vector, it blindly predicts straight-line paths, often resulting in projected collisions where humans would naturally curve their trajectories (Fig.~\ref{fig:cvm_baseline}). 

\begin{table*}[t]
    \centering
    \caption{Dynamic GraphSLAM Evaluation Metrics (Prediction Horizon = 20 steps, 2.0s). Lower is better for all metrics. \newline ATE (Absolute Trajectory Error), RPE (Relative Pose Error), SDE (Safety Distance Error)}
    \label{tab:metrics_evaluation}
    \begin{tabular}{lcccccc}
        \toprule
        \textbf{Method} & \textbf{Robot RMSE $\downarrow$} & \textbf{LM RMSE $\downarrow$} & \textbf{ATE $\downarrow$} & \textbf{RPE (Mean) $\downarrow$} & \textbf{SDE $\downarrow$} & \textbf{Pred. RMSE $\downarrow$} \\
        \midrule
        No $\mathcal{L}_{kin}$ & \textbf{0.388} & 0.485 & 0.186 & 0.039 & 0.041 & 1.338 \\
        CVM Baseline                   & \textbf{0.388} & 0.479 & 0.186 & 0.04 & 0.037 & \underline{1.21} \\
        Single-Agent (MLP)             & 0.4 & 0.5 & 0.185 & 0.039 & 0.048 & \textbf{1.173} \\
        Deterministic GAT (Ours)       & \textbf{0.388} & \textbf{0.477} & 0.184 & \textbf{0.039} & 0.038 & 1.355 \\
        \textbf{Stochastic GAT (Ours)} & \textbf{0.388} & \textbf{0.477} & \textbf{0.183} & 0.048 & \textbf{0.033} & \underline{1.21} \\
        \bottomrule
    \end{tabular}
\end{table*}


\subsection{Single-Agent Neural Prior (MLP)}

To address the limitations of linear extrapolation, we replaced the CVM factor with an independent single-agent neural predictor (MLP), trained on the coordinate history of individual pedestrians. Despite the non-linear capacity of the neural network, the MLP failed to learn collision avoidance behaviors, as it lacks contextual awareness of neighboring obstacles. Empirically, the MLP collapsed into a behavior nearly identical to the CVM (yielding a comparable Prediction RMSE of 1.173). It provided no tangible improvement to the retrospective SLAM state (Robot RMSE: 0.400, LM RMSE: 0.500) and generated the same linear, collision-prone predictions in the presence of crowds.
\begin{figure}[ht]
    \centering
    \includegraphics[width=\linewidth]{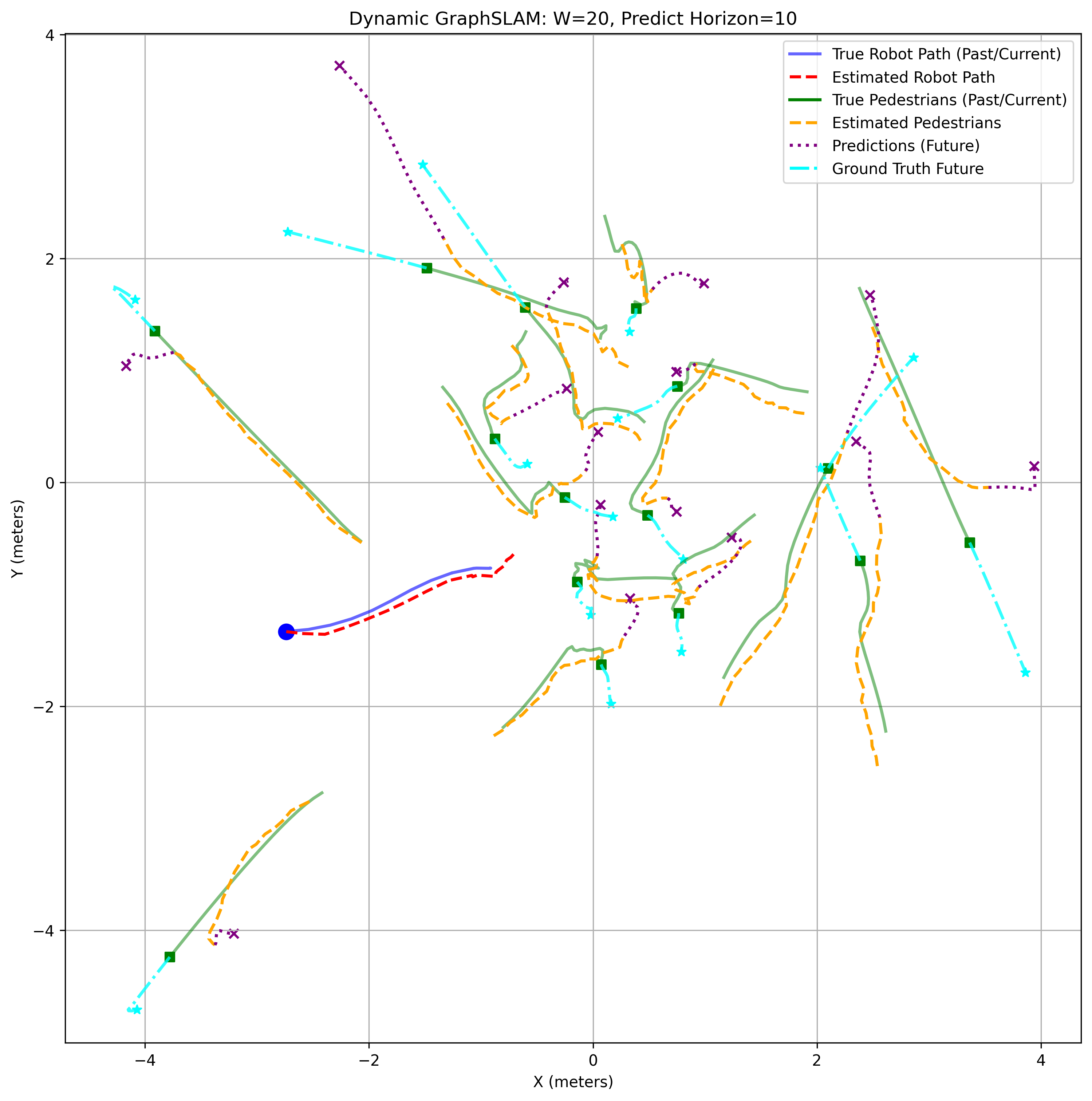}
    \caption{Single-Agent MLP Predictions. Lacking social context, the network degenerates into simple linear extrapolation, mirroring the CVM baseline and missing the ground-truth curvature.}
    \label{fig:mlp_baseline}
\end{figure}
\subsection{Multi-Agent Priors: Deterministic vs. Stochastic GAT}
\label{subsec:exp_gat}
Finally, we evaluate the multi-agent neural priors. The Deterministic GAT integrates social context via graph attention. While it attempts to model non-linear interactions, the deterministic formulation trained with an MSE loss suffers from a fundamental mathematical limitation: it converges to the conditional expected value of the training data. In partially observable pedestrian environments, interactions are inherently ambiguous (e.g., passing an oncoming person on the left vs. the right with a 50/50 probability). When forced to output a single deterministic trajectory, the GAT often regresses to the mean, generating an unphysical straight line (colliding directly into the obstacle) or producing jittery, overly aggressive trajectories. This "argmax problem" causes the deterministic model to diverge from reality, yielding a higher Prediction RMSE of 1.354.

\begin{figure}[htbp]
    \centering
    \begin{subfigure}{\linewidth}
        \includegraphics[width=\linewidth]{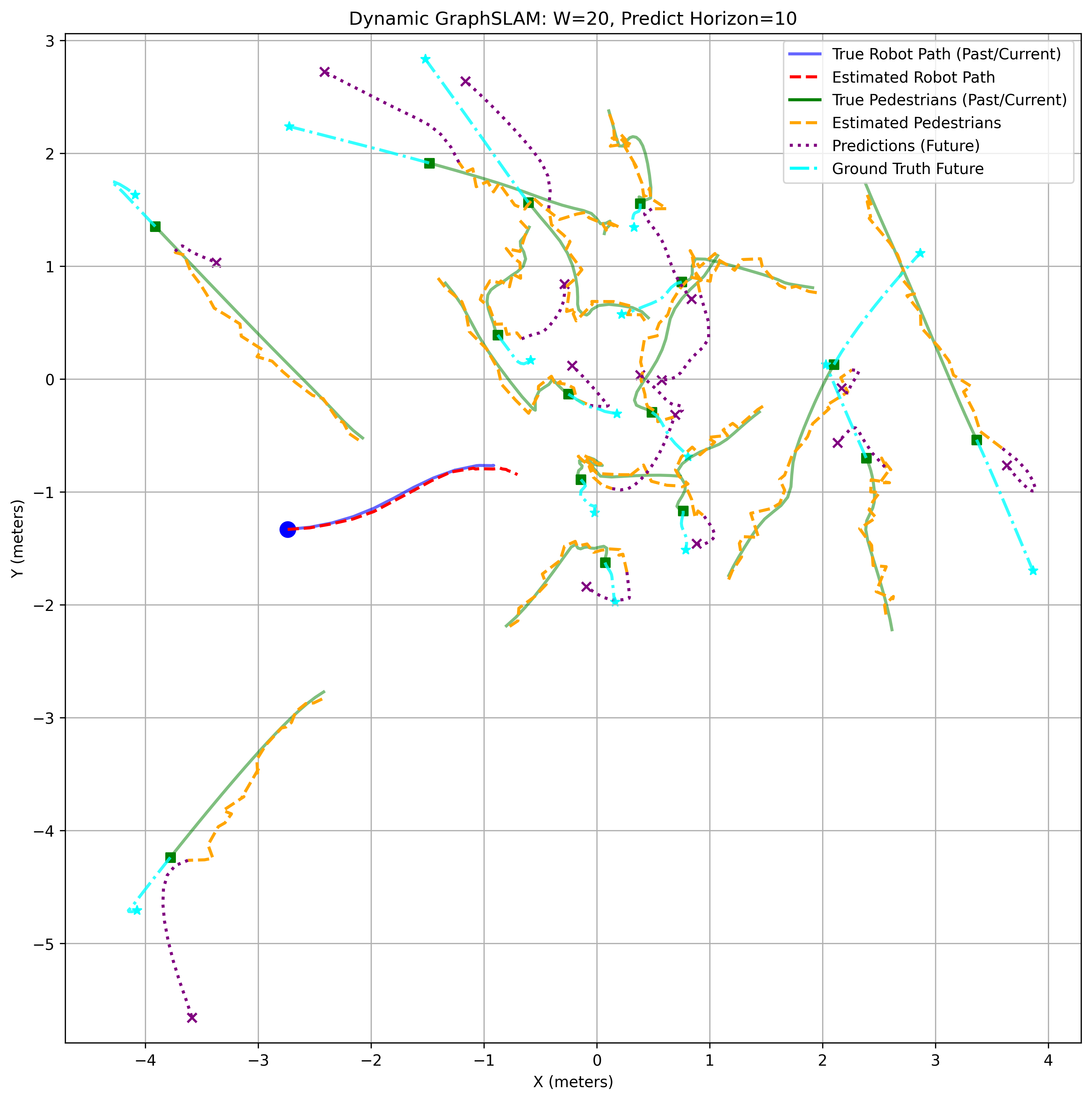}
        \caption{Deterministic GAT: The prediction attempts an avoidance maneuver but appears jittery and diverges from the ground truth.}
    \end{subfigure}
    \vspace{0.5cm} 
    \begin{subfigure}{\linewidth}
        \includegraphics[width=\linewidth]{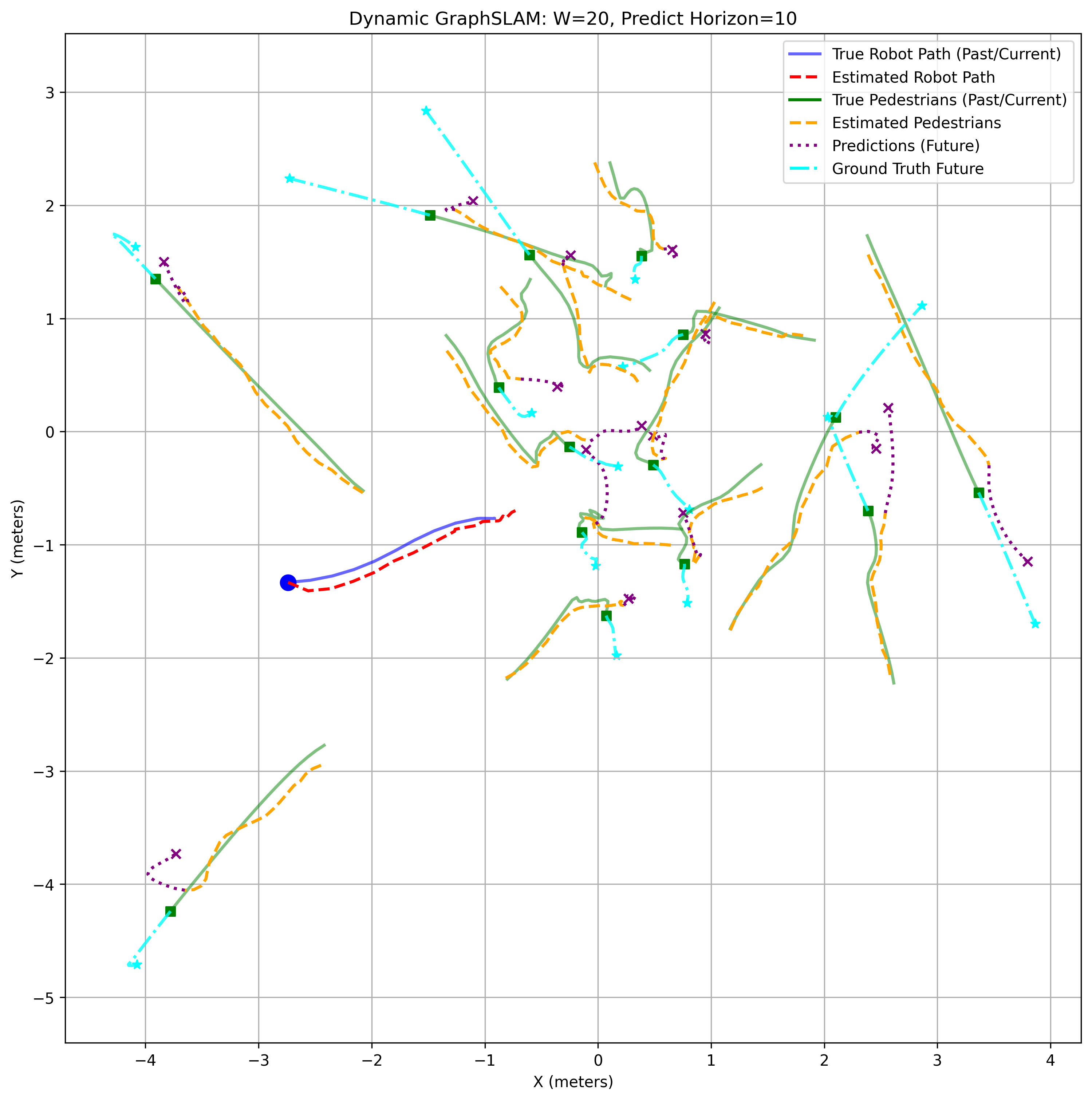}
        \caption{Stochastic GAT (Ours): By averaging multiple Monte Carlo rollouts and applying the Mahalanobis loss, the prediction becomes smooth and aligns perfectly with the true future.}
    \end{subfigure}
    \caption{Qualitative comparison of multi-agent prediction horizons within the dynamic SLAM graph.}
    \label{fig:gat_comparison}
\end{figure}

By transitioning to the \textbf{Stochastic GAT}, we overcome this limitation by treating the network as a lightweight \textit{World Model}. By querying the network using Monte Carlo perturbations, we generate multiple socially-aware rollouts that sample the true underlying multimodal distribution of human intents. As illustrated in Fig.~\ref{fig:gat_comparison}, even if the mathematical mean of these diverse rollouts happens to align with the Constant Velocity straight line (resulting in a similar Mean Prediction RMSE of 1.210), the Stochastic GAT correctly captures the \textit{epistemic uncertainty} of the interactions. 
This formulation allows the SLAM backend to compute a Mahalanobis distance based on the empirical covariance ($\Sigma_{kin}$) of the predictions. In open spaces, the rollouts converge, shrinking the covariance and strongly binding the SLAM graph to the prior. During complex interactions, the rollouts fan out, expanding the covariance ellipse. This dynamically relaxes the SLAM kinematic factor, preventing the optimizer from being misdirected by isolated bad predictions. More importantly, extracting this expanding uncertainty boundary provides downstream local planners (such as MPC or PPO) with a strict probabilistic safety envelope. Rather than blindly trusting a single maximum-likelihood trajectory, the control algorithm can optimize the robot's actions to safely avoid the entire distribution of potential non-linear human maneuvers.

\section{Conclusion}
\label{sec:conclusion}
In this work, we introduced a novel Dynamic GraphSLAM framework tightly coupled with socially-aware, generative motion priors. Our experiments on simulated highly-interactive environments demonstrate that traditional static SLAM approaches and Constant Velocity baselines fail to anticipate the complex, non-linear avoidance maneuvers characteristic of human crowds.

We showed that while single-agent neural networks (MLP) collapse to linear extrapolation due to a lack of spatial context, integrating a Graph Attention Network (GAT) effectively captures pedestrian social dynamics. Most crucially, we proved that framing the motion predictor as a stochastic World Model—generating multiple future hypotheses rather than a single deterministic output—fundamentally improves the system. By mapping the variance of these rollouts into a Mahalanobis distance within the SLAM factor graph, the system achieves dynamic stiffness, yielding highly accurate trajectory tracking (Robot RMSE of 0.39m) while explicitly quantifying interaction uncertainty.

Furthermore, this dynamic tracking system is designed to complement existing static SLAM pipelines. By running in parallel, our framework can refine the poses of moving and static landmarks and the robot simultaneously, effectively cleaning the dynamic noise that usually corrupts static mapping. Ultimately, the continuous extraction of multi-agent states and uncertainty covariance matrices provides an ideal input space for safety-critical control algorithms like Model Predictive Control (MPC), enabling anticipatory and collision-free robot navigation in densely populated human environments.

\section{Future Work}
\label{sec:future_work}
While the Stochastic GAT successfully models interaction uncertainty via Monte Carlo sampling, it relies on manually injected perturbations. In future work, we plan to replace this surrogate stochasticity with a fully generative \textbf{Flow Matching} architecture. Flow Matching will allow us to directly learn the continuous probability density vector fields of pedestrian trajectories, providing mathematically rigorous multi-modal generation without modifying the underlying coordinate history. 

Additionally, we aim to extend our experimental validation into the control domain. We will actively integrate the proposed SLAM pipeline with an advanced Model Predictive Controller (MPC) within the simulator to empirically measure collision rates and navigation efficiency. This will conclusively validate how the extracted covariance ellipses enhance the robot's real-time safety in reactive, multi-agent scenarios.

Another improvment lays down to a static landmarks. Our DynoSLAM can be combined with standart GraphSLAM with static landmarks, that will boost not only SLAM task performance, but the future predictions and control. 

Another interesting approach is training inside the simulation created by predictions of the trained GAT model -- it is the concept of World Models, where we can train an agent inside the virtual environments. There are works that show that training inside these virtual environment can boost control module performance.

\end{document}